\documentclass[letterpaper]{article} 
\usepackage{aaai2027}  
\usepackage[hyphens]{url}  
\usepackage{graphicx} 
\graphicspath{{Figures/}}
\usepackage{natbib}  
\usepackage{caption} 
\usepackage{latexsym}
\usepackage{booktabs}
\usepackage{multirow}
\usepackage{array}
\usepackage{amsmath}
\usepackage{amssymb}
\usepackage{fontawesome5}
\usepackage[most]{tcolorbox}
\usepackage{colortbl}
\usepackage{tabularx}
\newcommand{\R}{\mathbb{R}}

\newcommand{\softmax}{\operatorname{softmax}}
\newcommand{\rmsnorm}{\operatorname{RMSNorm}}
\newcommand{\layernorm}{\operatorname{LayerNorm}}

\newcommand{\JS}{\operatorname{JS}}
\newcommand{\Pool}{\operatorname{Pool}}

\newcommand{\Hop}{\mathrm{H}}
\newcommand{\MDS}{\mathrm{MDS}}

\newcommand{\bV}{\mathbf{V}}

\newcommand{\bq}{\mathbf{q}}

\newcommand{\balpha}{\boldsymbol{\alpha}}
\newcommand{\methodname}{LayerRoute}
\newcommand{\routername}{Layer Mixture Router}
\newcommand{\actionreread}{Action-State Reread}
\newcommand{\starvlapi}{StarVLA-\ensuremath{\pi}}

\newcommand{\Best}[1]{{\bfseries #1}}

\definecolor{abstractpanel}{RGB}{246,249,255}
\renewenvironment{abstract}{%
  \centerline{\bf Abstract}\vspace{0.45ex}%
  \begin{tcolorbox}[enhanced,frame hidden,colback=abstractpanel,arc=2mm,
    boxsep=0pt,left=7pt,right=7pt,top=7pt,bottom=7pt,
    before skip=0pt,after skip=5pt,fontupper=\small]
}{\end{tcolorbox}}
\copyrighttext{Email: \texttt{<lululzz0906@gmail.com>}\\[0.1ex]$^{*}$ Equal contribution. $^{\dagger}$ Corresponding authors.}

\title{LayerRoute: Action-Conditioned Mixture-of-Layers Routing for\\
Vision-Language-Action Policies}

\affiliations{}
\author{%
{\normalsize\bfseries Zheng Lu$^{1,2,*}$, Haoran Liao$^{1,*}$, Wanqi Zhong$^{1,*}$, Yunhe Ni$^{1}$, Lijie Wang$^{1,4}$, Xingjie Fan$^{1}$}\\[0.32em]
{\normalsize\bfseries Zhisheng Chen$^{5}$, Yantang Qu$^{1}$, Meijia Chen$^{6}$, Tianyu Xin$^{1}$, Zirui Song$^{3,\dagger}$, Yiming Li$^{1,\dagger}$}\\[0.53em]
{\normalfont\small $^{1}$ Tsinghua University \quad $^{2}$ Peking University \quad $^{3}$ MBZUAI}\\[0.18em]
{\normalfont\small $^{4}$ Zhejiang University \quad $^{5}$ Nanyang Technological University \quad $^{6}$ Rutgers University}
}

\begin{document}
\maketitle

\begin{abstract}
Vision-Language-Action (VLA) policies leverage pretrained vision-language
models (VLMs) to guide action generation for robot control.
VLMs provide hierarchical visual-semantic representations that evolve across layers, from
local visual geometry to abstract, language-aligned semantics; different
manipulation tasks may therefore require different mixtures of layer
representations. Meanwhile, the action module maintains intermediate
representations that evolve throughout action computation and may provide
useful information for subsequent decisions. However, existing VLA interfaces
offer limited flexibility in representation access: VLM information is exposed
through fixed layer assignments for each action layer, while intermediate
action states are only propagated implicitly through residual streams without
explicit reuse. We introduce \methodname{}, an action-conditioned
representation routing interface that enables adaptive access to VLM layers
and action representations. The Layer Mixture Router dynamically forms
mixtures of cached VLM representations, while Action-State Reread
reuses earlier action representations. Across diverse simulation and
real-world benchmarks, \methodname{} consistently improves \starvlapi{} and
\(\pi_{0.5}\), achieving up to 7.2 gains on LIBERO Long with only 0.31\% / 3.87\%
additional parameters. Ablation studies validate the benefit of action-conditioned layer
routing, while routing analyses reveal structured allocation patterns across
action layers and task settings.
\end{abstract}

\section{Introduction}
\label{sec:intro}

Vision-Language-Action (VLA) policies inherit visual
and semantic priors from pretrained vision--language models (VLMs)
and translate them into closed-loop robot behavior through
diverse action-generation architectures~\citep{zitkovich2023rt,
kim2024openvla,team2024octo,black2024pi_0,intelligence2025pi_}.
Despite their differences in action representation and generation,
these systems expose each action layer to predetermined
VLM layers, limiting the ability to adapt layer-wise VLM representation access to evolving
action computation and diverse manipulation requirements.

This limitation matters because VLM representations vary across layers. Analyses of ViTs, CLIP, and multimodal language models associate earlier states with local texture, geometry, and spatial patterns, and later states with abstract or language-aligned semantics~\citep{raghu2021vision,gandelsman2024interpreting,neo2025towards}. Robot manipulation requires different combinations of these cues across action tasks: grasp localization, gripper--object alignment, and contact-sensitive placement depend more strongly on fine-grained visual evidence, whereas instruction grounding, object-role disambiguation, and subgoal
selection rely more on abstract semantic representations. Since different
action tasks may require different mixtures of VLM layer representations, fixed layer
assignments cannot dynamically adapt representation access to evolving action requirements.

The limitation also arises within the action module. Under standard
residual propagation, earlier action states are progressively
transformed and integrated into the current representation, leaving
later blocks unable to revisit earlier intermediate action states.
This indirect access may hinder the retrieve of information needed for
later manipulation decisions, particularly in long-horizon manipulation. The policy is therefore constrained at two levels: fixed VLM-depth exposure limits access to external VLM representations,
while residual-only access limits reuse of internal action states.

\begin{figure*}[!t]
\centering
\includegraphics[width=\textwidth]{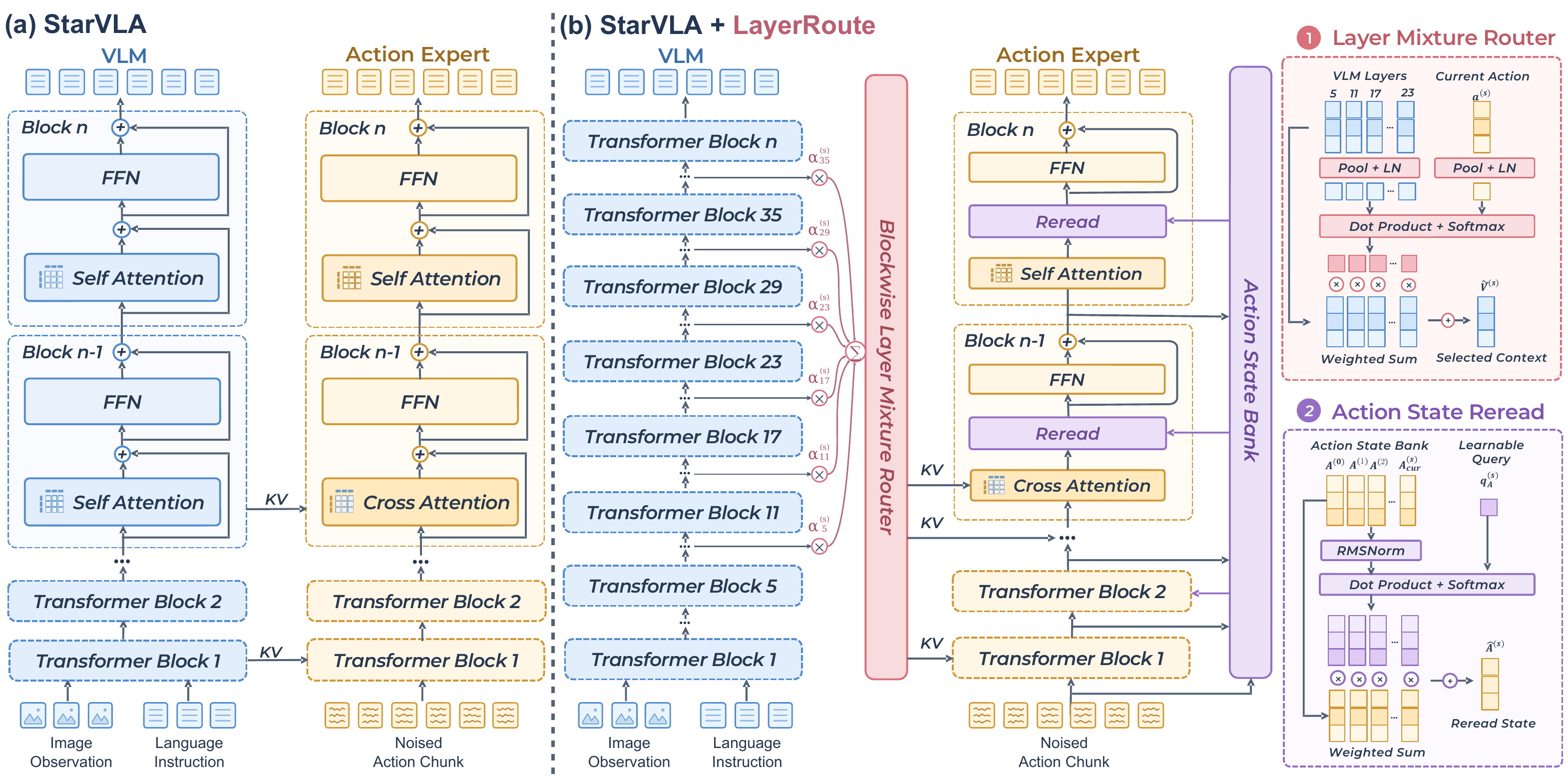}
\caption{Original standard StarVLA-$\pi$ versus StarVLA-$\pi$ with \methodname{}.
The original interface uses fixed VLM context and residual-only action-state
access, while \methodname{}
enables action-conditioned layer-wise routing and earlier action-state
reuse.}
\label{fig:teaser}
\end{figure*}

We argue that enabling adaptive access to VLM layers and
earlier action states can improve control beyond fixed representation
interfaces. Motivated by this view, we introduce \methodname{}, a
representation routing interface that selects VLM representations
across depth and retrieves earlier action states in the action module
(Figure~\ref{fig:teaser}). The \routername{} conditions layer-wise VLM representation access on the current action state, while \actionreread{} enables
action blocks to reuse earlier action representations.

We train \methodname{} on two representative VLA
backbones, \starvlapi{} and \(\pi_{0.5}\). With modest overhead,
\methodname{} improves LIBERO performance, increasing Long success rates by
7.2 and 3.6 and Avg\(_4\) by 2.3 and 1.3 for
\starvlapi{} and \(\pi_{0.5}\), respectively. These improvements transfer to SimplerEnv and RoboCasa-GR1, where
\methodname{} improves performance across robot settings, with gains of up to
5.6 points on SimplerEnv WidowX. The improvements generalize across VLM backbones, including
Qwen3-VL, Qwen2.5-VL~\citep{bai2025qwen3},
MiMo-Embodied, and
Cosmos-Reason2-2B~\citep{nvidia2026cosmosreason2}. Routing ablations 
validate the mechanism, while routing analyses reveal layer
allocation patterns across action layers and task settings.

\section{Related Work}
\label{sec:related}

\paragraph{VLA Action Generation and Representation Interfaces.}
VLA and generalist robot policies differ in how actions are decoded.
RT-2~\citep{zitkovich2023rt} and
OpenVLA~\citep{kim2024openvla} map robot controls to discrete
language-model tokens, Octo~\citep{team2024octo} uses a diffusion action
head, and \(\pi_0\)/\(\pi_{0.5}\)~\citep{black2024pi_0,
intelligence2025pi_} use flow-matching action experts, with
\(\pi_{0.5}\) additionally incorporating tokenized-action pretraining.
StarVLA~\citep{community2026starvla} adopts a modular VLM--action architecture,
while Knowledge Insulation~\citep{driess2026knowledge} studies the
interaction between continuous action experts and VLM training.
These works primarily study action representation, decoding, or
VLM--expert coupling. We instead study how action layers access
pretrained VLM representations across depth.

\paragraph{Adaptive Representation Access and Routing.}
OTTER~\citep{huang2025otter} selects instruction-relevant visual
features, CogVLA~\citep{li2026cogvla} aggregates and prunes visual
tokens, VLA-Cache~\citep{xu2026vla} reuses stable visual-token
key--value states across frames, and AVA-VLA~\citep{xiao2025ava}
uses execution history to reweight current visual tokens.
FedVLA~\citep{miao2025fedvla} studies token and expert routing in a
federated setting. SpatialVLA~\citep{qu2025spatialvla} introduces
Ego3D position encoding and adaptive action grids. MiMo-Embodied~\citep{hao2025mimo} targets
affordance prediction, task planning, and spatial understanding,
while Causal Planner~\citep{lu2026token} emphasizes
physically grounded next-state reasoning. HAMLET and MemoryVLA~\citep{koo2025hamlet,shi2025memoryvla} aggregate
information across environment timesteps. These works address visual
feature selection, embodied reasoning, and cross-timestep context
aggregation, whereas \methodname{} focuses on depth-aware access to VLM
states and direct reuse of action states during action computation.

\paragraph{Layerwise Representations and Cross-Layer Reuse.}
Prior studies reveal layerwise structures in ViTs, CLIP,
VLMs, and language models by analyzing representation evolution,
attention patterns, visual-token processing, and intermediate
predictions~\citep{raghu2021vision,gandelsman2024interpreting,neo2025towards,
belrose2023eliciting}. These analyses establish layer depth as a representation axis, but do not address which VLM representations
should be exposed to action computation. Beyond representation analysis, prior architectures modify information flow:
Highway Networks and stochastic depth regulate inter-layer propagation
or execution~\citep{srivastava2015training,huang2016deep};
DenseFormer and Attention Residuals aggregate or reuse earlier hidden
states~\citep{pagliardini2024denseformer,team2026attention};
Mixture-of-Depths, DeeR-VLA, and MoLe-VLA adapt computation
depth~\citep{raposo2024mixture,yue2024deer,zhang2026mole}, while
Hyper-Connections learn cross-depth
pathways~\citep{zhu2025hyper}. These methods alter state
propagation or computation allocation within a model, whereas
\methodname{} adapts representation access across the VLM--action
boundary and enables token-wise reuse of earlier action states.
Table~\ref{tab:positioning} compares these approaches from
perspective of representation access.

\begin{table*}[!tbp]
\centering
\captionsetup{labelfont=bf,labelsep=period,justification=raggedright,singlelinecheck=false,skip=5pt}
\caption{Representative interfaces for adaptive representation access and state
reuse~\citep{pagliardini2024denseformer,raposo2024mixture,
team2026attention}. Existing approaches modify VLM exposure, computation
depth, or hidden-state aggregation, while \methodname{} enables
action-conditioned routing over VLM layers together with explicit
action-state rereading.}
\label{tab:positioning}
\small
\definecolor{tableoneaccent}{HTML}{005A9C}
\definecolor{tableonehead}{RGB}{235,243,249}
\definecolor{tableonestripe}{RGB}{248,250,252}
\definecolor{tableoneours}{RGB}{225,239,249}
\setlength{\tabcolsep}{4pt}
\renewcommand{\arraystretch}{1.28}

\begin{tabularx}{\textwidth}{@{}>{\raggedright\arraybackslash}p{0.20\textwidth}>{\raggedright\arraybackslash}p{0.23\textwidth}>{\centering\arraybackslash}p{0.18\textwidth}>{\centering\arraybackslash}p{0.14\textwidth}>{\raggedright\arraybackslash}X@{}}
\toprule
\rowcolor{tableonehead}
\textbf{Interface}
&
\textbf{Representation}
&
\textbf{Action-Cond. Routing}
&
\textbf{State Reuse}
&
\textbf{Mechanism}
\\
\midrule

Final-layer readout
& final VLM representation
& No
& None
& fixed readout
\\

\rowcolor{tableonestripe}
Fixed layer mapping
& predefined VLM layers
& No
& None
& fixed cross-attention
\\

Prefix/suffix joint attention
& same-layer VLM stream
& No
& None
& joint attention
\\

\rowcolor{tableonestripe}
Token/expert routing
& visual tokens / experts
& No
& None
& token selection
\\

\addlinespace[0.18em]
\midrule
\addlinespace[0.18em]

DenseFormer
& current/prior hidden states
& N/A
& backbone
& hidden aggregation
\\

\rowcolor{tableonestripe}
Computation depth
& token/layer execution
& No
& None
& depth allocation
\\

Attention Residuals
& previous layer/block states
& N/A
& backbone
& residual aggregation
\\

\addlinespace[0.18em]
\midrule
\addlinespace[0.18em]
\rowcolor{tableoneours}
\textcolor{tableoneaccent}{\textbf{LayerRoute}}
& \textbf{cached VLM/action states}
& \textbf{Yes}
& \textbf{action module}
& \textbf{layer routing + reread}
\\

\bottomrule
\end{tabularx}

\end{table*}

\section{Background and Formulation}
\label{sec:background}

\paragraph{layer-wise VLM representation access.}
A pretrained VLM produces a hierarchy of hidden states
\(\bV_n\in\R^{B\times T_v\times D_v}\) at boundaries
\(n\in\{0,\ldots,N\}\), where \(n=0\) denotes the embedded input.
We retain these states as a multi-layer representation cache. Since
intermediate states preserve token correspondence across depth, they can
be directly combined along the depth dimension. At action layer \(s\), let
\(\mathbf{a}^{(s)}\in\R^{B\times T_a\times D_a}\) denote the current
action state, and let
\(\mathcal{C}_s\subseteq\{0,\ldots,N\}\) specify the VLM layers
available at that layer. We formulate layer routing as a
readout problem: a static interface accesses fixed VLM layers, whereas
our action-conditioned readout adapts this access based on the current
action state.
\[
\begin{gathered}
\widetilde{\bV}_{\mathrm{static}}^{(s)}
=
\Phi_s^{\mathrm{static}}
\bigl(\{\bV_n\}_{n\in\mathcal{C}_s}\bigr),\\
\widetilde{\bV}^{(s)}
=
\Phi_s
\bigl(
\{\bV_n\}_{n\in\mathcal{C}_s},
\mathbf{a}^{(s)}
\bigr).
\end{gathered}
\]
Here, \(\Phi_s^{\mathrm{static}}\) and \(\Phi_s\) denote the readout
functions at action layer \(s\). The static readout does not depend on
\(\mathbf{a}^{(s)}\), whereas the action-conditioned readout adapts the
VLM-layer combination according to the current action state.

\paragraph{Action-State Access.}
Representation access is also constrained within the action module.
Under residual propagation, earlier action states are progressively
transformed and integrated into the current representation. Later layers can therefore access earlier action states only through the residual stream rather than directly revisiting intermediate states. \methodname{} addresses
this limitation with \actionreread{}, which reuses earlier action states from the action module. Together with the
action-conditioned layer-wise routing introduced above, it forms a
representation routing interface for adaptive access to both VLM and
action representations.

\paragraph{Architectural Instantiations.}
We then apply this representation-access formulation to two representative VLA architectures. \starvlapi{}~\citep{community2026starvla} couples Qwen3-VL~\citep{bai2025qwen3} to an action DiT through cross-attention, whereas \(\pi_{0.5}\)~\citep{intelligence2025pi_} combines PaliGemma~\citep{beyer2024paligemma} with a flow-matching action expert. These architectures allow us to examine representation access under different VLA designs.

\section{Method}
\label{sec:method}

At action block \(s\), Action-Conditioned Layer Mixture Router pools current action state and cached VLM representations to predict a sample-wise distribution over VLM depth. It mixes same-position VLM tokens using this distribution and uses the result as the cross-attention context. After attention, Action-State Reread mixes the same action-token position across selected earlier action blocks. The resulting state is passed to the FFN and subsequent action blocks.

\subsection{Action-Conditioned Layer Mixture Router}
\label{sec:method-router}

Given the current action state and candidate VLM states, the router
computes an action-conditioned distribution over accessible VLM layers.
At a selected action layer \(s\), let
\(\mathcal{C}_s\subseteq\{0,\ldots,N\}\) denote the indices of the
available VLM states. The routed VLM representation is denoted by
\(\widetilde{\bV}^{(s)}\).

We first obtain compact summaries of the current action state and each
candidate VLM state by pooling over valid token positions:
\(\mathbf{p}_a^{(s)}
=
\Pool\!\left(\mathbf{a}^{(s)}\right)\),
\(\mathbf{p}_{V,n}
=
\Pool\!\left(\bV_n\right)\), \(n\in\mathcal{C}_s\).
The action and VLM summaries are separately normalized and projected
with layer-specific mappings into a shared router space of dimension
\(d_{\mathrm{sel}}=256\):
\[
\begin{gathered}
\bq^{(s)}
=
\mathbf{W}_{q,d}^{(s)}
\layernorm_{a}^{(s)}
\!\left(\mathbf{p}_{a}^{(s)}\right),\\
\mathbf{k}_{n}^{(s)}
=
\mathbf{W}_{k,d}^{(s)}
\layernorm_{V}^{(s)}
\!\left(\mathbf{p}_{V,n}\right),\\
\bq^{(s)},\mathbf{k}_{n}^{(s)}
\in\R^{d_{\mathrm{sel}}},
\qquad n\in\mathcal{C}_s.
\end{gathered}
\]
The scaled dot-product similarity produces routing logits, which are
normalized into depth weights over the available VLM states:
\[
\begin{gathered}
\ell_{n}^{(s)}
=
\frac{
\left\langle
\bq^{(s)},\mathbf{k}_{n}^{(s)}
\right\rangle
}{
\sqrt{d_{\mathrm{sel}}}
},
\qquad n\in\mathcal{C}_s,\\
\alpha_{n}^{(s)}
=
\frac{
\exp\!\left(\ell_n^{(s)}\right)
}{
\sum_{m\in\mathcal{C}_s}
\exp\!\left(\ell_m^{(s)}\right)
},
\qquad n\in\mathcal{C}_s.
\end{gathered}
\]
The routed representation is obtained by aggregating candidate VLM
states with these depth weights:
\mbox{\(\widetilde{\bV}^{(s)}
=
\sum_{n\in\mathcal{C}_s}
\alpha_n^{(s)}\,\bV_n\)}.

The router predicts a single depth distribution for each selected
action layer, shared across all VLM token positions.

\subsection{Internal Action-State Reread}
\label{sec:method-action-reread}

\actionreread{} enables each action token to reuse intermediate action
states from earlier layers. At a selected
action layer \(s\), the accessible source states are indexed by
\(\mathcal{I}_s=\{0,j_1,\ldots,j_k,\mathrm{cur}\}\),
\(\mathbf{A}_{\mathrm{cur}}=\mathbf{a}^{(s)}\),
where \(\mathbf{A}_0\) denotes the initial action state,
\(\mathbf{A}_{j_1},\ldots,\mathbf{A}_{j_k}\) denote states retained from
earlier action layers, and \(\mathbf{A}_{\mathrm{cur}}\) denotes the
current state.

Each selected action layer maintains a learned query
\(\mathbf{q}_A^{(s)}\in\R^{D_a}\), shared across all action tokens in
that layer. For token \(t\), the learned query
\(\mathbf{q}_A^{(s)}\) measures its compatibility with the token
representation \(\mathbf{A}_{i,t}\) from each accessible source state
\(i\in\mathcal{I}_s\):
\[
\begin{gathered}
\eta_{t,i}^{(s)}
=
\left(\mathbf{q}_{A}^{(s)}\right)^\top
\rmsnorm\!\left(\mathbf{A}_{i,t}\right),
\qquad i\in\mathcal{I}_s,\\
\rho_{t,i}^{(s)}
=
\frac{\exp\!\left(\eta_{t,i}^{(s)}\right)}
{\sum_{r\in\mathcal{I}_s}
 \exp\!\left(\eta_{t,r}^{(s)}\right)},
\qquad i\in\mathcal{I}_s.
\end{gathered}
\]

The reread state for token \(t\) is obtained by aggregating source
states with these weights:
\(\widehat{\mathbf{a}}_t^{(s)}
=
\sum_{i\in\mathcal{I}_s}
\rho_{t,i}^{(s)}\,\mathbf{A}_{i,t}\).
Stacking token-level outputs yields the reread action state
\(\widehat{\mathbf{a}}^{(s)}\). Although the query is shared within
each layer, source states vary across token positions, resulting in
token-specific weights \(\rho_{t,i}^{(s)}\). This differs from the VLM
router, which predicts a single depth distribution for each sample
and selected action layer that is shared across all VLM token
positions.

\subsection{Integration into \starvlapi{}}
\label{sec:method-placement}

In \starvlapi{}, the action module is a 36-layer DiT with 18
cross-attention layers and 18 self-attention layers. We retain
representations from six predefined VLM depths and expose them to all
cross-attention layers, allowing each layer to construct its depth
mixture conditioned on the current action state rather than using a
fixed VLM representation. The six-depth choice is evaluated in Supplementary Table~2.

At cross-attention layer \(s\), the router uses
\(\mathbf{a}^{(s)}\) to compute routing weights over the retained VLM
states and obtain \(\widetilde{\bV}^{(s)}\), which replaces the encoder
context of the original interface:
\(\mathbf{Z}^{(s)}
=
\widetilde{\bV}^{(s)}\).
The cross-attention takes \(\mathbf{a}^{(s)}\) as the query and
\(\mathbf{Z}^{(s)}\) as the key--value context, producing
\(\mathbf{a}_{\mathrm{attn}}^{(s)}\). Applied at all 18
cross-attention layers, the router enables action-conditioned
layer-wise routing throughout the DiT.

We apply \actionreread{} to all 36 DiT blocks, including the 18
self-attention blocks. At each block, the attention output serves as
the current action-state source. For cross-attention blocks,
\(\mathbf{A}_{\mathrm{cur}}
=
\mathbf{a}_{\mathrm{attn}}^{(s)}\).
For self-attention blocks, \(\mathbf{A}_{\mathrm{cur}}\) is obtained
analogously from the post-self-attention state. The current state is
then recombined with earlier action representations to produce
\(\widehat{\mathbf{a}}^{(s)}\), which is passed to the feed-forward
network. Thus, the router provides adaptive layer-wise VLM representation access at the
18 cross-attention layers, while \actionreread{} enables reuse of
earlier action states throughout the 36 DiT blocks.

We extend the same representation routing interface \methodname{} to \(\pi_{0.5}\), with its
backbone-specific implementation detailed in Supplementary Table 1.

\section{Experiments}
\label{sec:experiments}

We evaluate \methodname{} on two VLA backbones across three simulation
benchmarks. We further study its generality across VLM backbones,
examine the roles of the two representation reads through ablations,
and analyze the learned structured routing behavior through layer-routing
analysis.

\subsection{Setup}
\label{sec:exp-setup}

We evaluate \starvlapi{}~\citep{community2026starvla} and
\(\pi_{0.5}\)~\citep{intelligence2025pi_} on three simulation
benchmarks~\citep{liu2023libero,li2024evaluating,
nasiriany2024robocasa}. LIBERO evaluates diverse tabletop
manipulation across spatial, object-conditioned, goal-conditioned, and
long-horizon tasks. SimplerEnv measures robustness under visual and
environmental variations across different robot platforms, while
RoboCasa-GR1 extends evaluation to household manipulation with a
humanoid embodiment. Together, these benchmarks assess representation
access under diverse task structures, environments, and
control demands across manipulation scenarios.

All models are trained on eight NVIDIA H200 GPUs with a per-device
batch size of 32. The pretrained VLM, action module, router, and
action-state reread are jointly fine-tuned. For \starvlapi{}, the
multi-depth representation cache retains six Qwen3-VL hidden states from
layers \([5,11,17,23,29,35]\). A \(256\)-dimensional router operates at
all 18 cross-attention layers of the 36-layer action DiT. Training uses
AdamW with backbone-specific learning rates, \(10^{-8}\) weight decay and
5K-step warmup; for \starvlapi{}, the base parameters use
\(2.5\times10^{-5}\), while the action and routing parameters use
\(10^{-4}\). Dataset, training, and benchmark-specific evaluation settings
are detailed under \emph{Experimental Setup} and \emph{Cross-benchmark
execution} in the Supplementary.

\subsection{LIBERO Evaluation across Backbones}
\label{sec:exp-starvla}

Table~\ref{tab:libero-starvla} reports the performance of
\methodname{} on \starvlapi{} and \(\pi_{0.5}\). Across both
backbones, \methodname{} improves LIBERO performance. Specifically,
\starvlapi{} gains 2.3 and 7.2 percentage points on Avg\(_4\) and Long,
while \(\pi_{0.5}\) gains 1.3 and 3.6 points, respectively. These
results demonstrate that the representation-access mechanism
generalizes across VLA architectures with distinct action formulations. The Long-suite
improvements further suggest that adaptive layer-wise VLM representation access and
earlier action-state reuse are beneficial for long-horizon
manipulation.

\subsection{Cross-Benchmark Evaluation}
\label{sec:exp-cross-benchmark}

We further evaluate \methodname{} on SimplerEnv and RoboCasa-GR1.
Across both benchmarks, \methodname{} consistently improves the
performance of \starvlapi{} and \(\pi_{0.5}\). On SimplerEnv,
\starvlapi{} and \(\pi_{0.5}\) achieve gains of 1.6/2.5 and 2.1/5.6
points on Overall/WidowX success, respectively
(Table~\ref{tab:simpler}). On RoboCasa-GR1, \methodname{} improves
macro-average success by 1.2 and 5.5 points for \starvlapi{} and
\(\pi_{0.5}\), respectively (Table~\ref{tab:robocasa-gr1}). These
improvements are achieved with modest parameter overheads of 0.31\%
and 3.87\% and latency increases of 10.03\% and 25.04\%, yielding a
favorable accuracy--efficiency trade-off. RoboCasa-GR1 task-level results and SimplerEnv task/task-family breakdowns are reported in Supplementary Tables~4, 7, and 8. Overall, these results show that LayerRoute remains effective across
diverse manipulation settings, including different tasks, environments,
and robot embodiments.

\subsection{Sensitivity to the VLM Backbone}
\label{sec:exp-vlm-sensitivity}

\begin{table}[!t]
\centering
{\small
\setlength{\tabcolsep}{2pt}
\begin{tabular*}{\linewidth}{@{}l@{\extracolsep{\fill}}ccccc@{}}
\toprule
\multirow{2}{*}{Model} & \multicolumn{5}{c}{SR (\%) \(\uparrow\)} \\
      & Spatial & Object & Goal & Long & Avg\(_4\) \\
\midrule
\starvlapi{} 
& 98.8 & \Best{99.6} & 95.8 & 88.4 & 95.7 \\

\starvlapi{}+\methodname{} 
& 98.6 & 99.2 & 98.4 & \Best{95.6} & 98.0 \\

\(\pi_{0.5}\) 
& 98.8 & 98.2 & 98.0 & 92.4 & 96.9 \\

\(\pi_{0.5}\)+\methodname{} 
& \Best{99.0} & 98.4 & \Best{99.2} & \Best{96.0} & \Best{98.2} \\
\bottomrule
\end{tabular*}
}
\caption{LIBERO success rate. Each suite contains 500 evaluation
episodes; Avg\(_4\) is the unweighted mean across suites.
\starvlapi{} uses Qwen3-VL~\citep{bai2025qwen3}. Bold indicates the best
result in each column throughout the paper.}
\label{tab:libero-starvla}

{\small
\setlength{\tabcolsep}{0pt}
\begin{tabular*}{\linewidth}{@{}l@{\extracolsep{\fill}}cccc@{}}
\toprule
\multirow{2}{*}{Model}
      & VM
      & VA
      & GR Avg.
      & WidowX \\
      & SR (\%) \(\uparrow\)
      & SR (\%) \(\uparrow\)
      & SR (\%) \(\uparrow\)
      & SR (\%) \(\uparrow\) \\
\midrule
\starvlapi{}
& 75.1 & 69.0 & 72.1 & 60.8 \\

\starvlapi{}+\methodname{}
& \Best{76.8} & \Best{70.6}
& \Best{73.7} & \Best{63.3} \\

\(\pi_{0.5}\)
& 73.7 & 67.5 & 70.6 & 57.1 \\

\(\pi_{0.5}\)+\methodname{}
& 76.1 & 69.3 & 72.7 & 62.7 \\
\bottomrule
\end{tabular*}
}
\caption{SimplerEnv success rate. VM/VA denote Google Robot Visual
Matching/Variant Aggregation; GR Avg. averages VM/VA, and WidowX
is reported separately.}
\label{tab:simpler}

{\small
\setlength{\tabcolsep}{1.6pt}
\begin{tabular*}{\linewidth}{@{}>{\raggedright\arraybackslash}p{0.45\linewidth}@{\extracolsep{\fill}}>{\centering\arraybackslash}p{0.15\linewidth}>{\centering\arraybackslash}p{0.17\linewidth}>{\centering\arraybackslash}p{0.18\linewidth}@{}}
\toprule
\multirow{2}{*}{Model} & GR1 Avg. & \multicolumn{2}{c}{Overhead (\%) \(\downarrow\)} \\
      & SR (\%) \(\uparrow\) & Params & Latency \\
\midrule

\starvlapi{}
& 43.9 & \Best{0} & \Best{0} \\

\starvlapi{}+\methodname{}
& \Best{45.1} & 0.31\% & +10.03\% \\

\(\pi_{0.5}\)
& 37.0 & \Best{0} & \Best{0} \\

\(\pi_{0.5}\)+\methodname{}
& 42.5 & 3.87\% & +25.04\% \\

\bottomrule
\end{tabular*}
}
\caption{Macro-average success rate over 24 RoboCasa-GR1 tasks. Added parameters and latency increases are relative to the corresponding backbone.}
\label{tab:robocasa-gr1}

{\small
\setlength{\tabcolsep}{2.0pt}
\begin{tabular*}{\linewidth}{@{}l@{\extracolsep{\fill}}ccc@{}}
\toprule
\multirow{2}{*}{VLM backbone} & Original interface & \methodname{} & Gain \\
             & Avg\(_4\) SR (\%) \(\uparrow\) & Avg\(_4\) SR (\%) \(\uparrow\) & (pp) \(\uparrow\) \\
\midrule
Qwen3-VL & \Best{95.7} & 98.0 & +2.3 \\
Qwen2.5-VL & 95.0 & 96.2 & +1.2 \\
MiMo-Embodied & 94.9 & \Best{98.2} & \Best{+3.3} \\
Cosmos-Reason2-2B & 95.4 & 97.4 & +2.0 \\
\bottomrule
\end{tabular*}
}
\caption{VLM-backbone sensitivity on LIBERO Avg\(_4\). Gain (pp) is
\methodname{} minus the original interface.}
\label{tab:vlm-sensitivity}

\begin{center}
\centering
{\small
\setlength{\tabcolsep}{1.0pt}
\begin{tabular*}{\linewidth}{@{}l@{\extracolsep{\fill}}ccc@{}}
\toprule
\multirow{2}{*}{Backbone} & \multirow{2}{*}{Variant} &
Avg\(_4\) &
Long \\
& &
SR (\%) \(\uparrow\) &
SR (\%) \(\uparrow\) \\
\midrule
\starvlapi{} & static interface
& 95.7 & 88.4 \\
\starvlapi{} & VLM router only
& 96.4 & 92.4 \\
\starvlapi{} & action-state reread only
& 95.9 & 93.8 \\
\starvlapi{} & VLM router + action-state reread
& \Best{98.0} & \Best{95.6} \\
\midrule
\(\pi_{0.5}\) & static interface
& 96.9 & 92.4 \\
\(\pi_{0.5}\) & VLM router only
& 97.3 & 93.6 \\
\(\pi_{0.5}\) & action-state reread only
& 95.5 & 94.2 \\
\(\pi_{0.5}\) & VLM router + action-state reread
& \Best{98.2} & \Best{96.0} \\
\bottomrule
\end{tabular*}
}
\caption{Component ablations on LIBERO. Avg\(_4\) denotes the
average SR (\%) across four suites.}
\label{tab:variants-results}
\end{center}
\end{table}

To assess the sensitivity of \methodname{} to the VLM backbone, we keep
the \starvlapi{} action architecture and training configuration fixed
while varying only the VLM backbone that provides the multi-layer representations.

As shown in Table~\ref{tab:vlm-sensitivity}, \methodname{} improves
Avg\(_4\) relative to the corresponding static interface for all four
VLM backbones, with gains ranging from 1.2 to 3.3 percentage points.
These consistent improvements across VLM backbones demonstrate that
adaptive representation access remains effective across different VLM
choices.

\subsection{Component Ablations}
\label{sec:exp-ablation}

Table~\ref{tab:variants-results} presents controlled ablations of the two routing mechanisms. We compare the static interface, VLM router only, action-state reread only, and complete routing interface.

Across both backbones, the complete routing interface achieves the best
Avg\(_4\) and Long performance. Compared with the best-performing single-read
variant, it improves Avg\(_4\)/Long by 1.6/1.8 points on
\starvlapi{} and 0.9/1.8 points on \(\pi_{0.5}\), respectively. These results
demonstrate that adaptive access to both VLM layer representations and
intermediate action states provides complementary benefits.

\section{Layer-Routing Analysis}
\label{sec:analysis}

We further examine whether the Layer Mixture Router benefits from action-guided
layer-wise routing by comparing it with static, uniform, and fixed-depth
alternatives. We also analyze how the learned VLM-depth distributions
evolve during training and vary across action layers and task suites.

\begin{table}[!b]
\centering
{\small
\setlength{\tabcolsep}{1.2pt}
\begin{tabular*}{\linewidth}{@{}l@{\extracolsep{\fill}}ccc@{}}
\toprule
Routing configuration
& Avg\(_4\) SR \(\uparrow\)
& Long SR \(\uparrow\)
& \(\Delta_{\mathrm{Long}}\) \\
\midrule
original static interface
& 95.7 & 88.4 & 0.0 \\
\midrule
uniform route
& 94.2 & 92.2 & +3.8 \\
forced shallow route
& 96.3 & 91.2 & +2.8 \\
forced middle route
& 95.9 & 93.0 & +4.6 \\
forced deep route
& 96.4 & 94.2 & +5.8 \\
action-independent query
& 97.2 & 94.8 & +6.4 \\
current-state route
& \Best{98.0} & \Best{95.6} & \Best{+7.2} \\
\bottomrule
\end{tabular*}
}
\caption{\starvlapi{} routing-policy controls. All routed variants differ only in the layer-wise routing rule. \(\Delta_{\mathrm{Long}}\) is measured relative to the original static
interface.}
\label{tab:route-policy-ablation}
\end{table}

\subsection{Routing-Policy Controls}
\label{sec:analysis-specificity}

Table~\ref{tab:route-policy-ablation} examines whether conditioning layer-wise routing on the current action state improves performance relative to uniform, fixed-depth, and action-independent routing alternatives. In the
action-independent control, the pooled action representation
\(\mathbf{p}_a^{(s)}\) is replaced with a fixed nonzero vector
\(\mathbf{c}_0\):
\(\bq_{\mathrm{ind}}^{(s)}
=
\mathbf{W}_{q,d}^{(s)}
\layernorm_a^{(s)}\!\left(\mathbf{c}_0\right)\).
The resulting query remains learned and layer-specific, but the routing
weights no longer depend on the current action state. The action-guided
route achieves 98.0 Avg\(_4\) and 95.6 Long, exceeding the
action-independent control by 0.8 points on both metrics, the strongest
fixed-depth route by 1.6/1.4 points, and uniform routing by 3.8/3.4
points. These comparisons show that conditioning
layer-wise routing on the current action representation provides an
effective routing signal.

\subsection{Routing Diagnostics}
\label{sec:analysis-diagnostics}

\begin{figure}[!tb]
\centering
\includegraphics[width=\columnwidth]{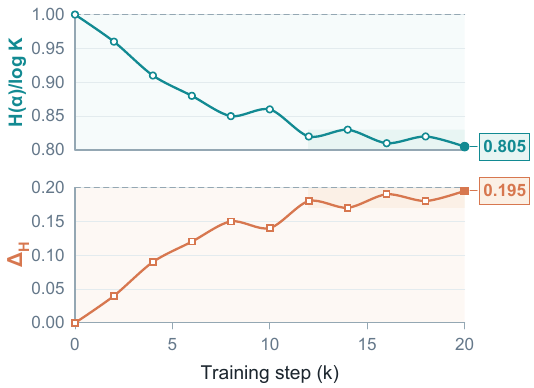}
\caption{Routing selectivity over training. Normalized entropy
\(H(\balpha)/\log K\) and entropy deficit \(\Delta_{\Hop}\) are complementary
measures: entropy decreases as \(\Delta_{\Hop}\) increases. Shaded regions
indicate routing variation in late-training window.}
\label{fig:entropy}
\end{figure}

\begin{figure}[!b]
\centering
\includegraphics[width=\columnwidth]{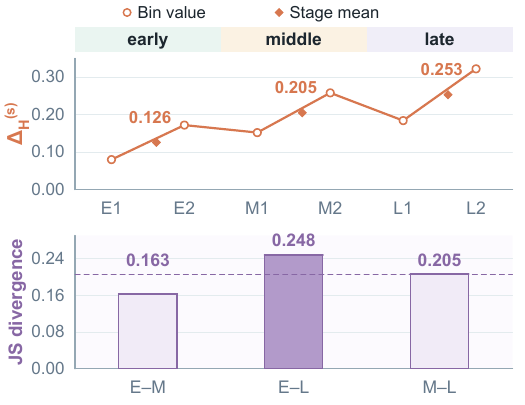}
\caption{Routing selectivity across action-layer groups. Top:
bin-level normalized entropy deficit \(\Delta_{\Hop}\) (circles) and
early/middle/late means of 0.126/0.205/0.253 (diamonds). Bottom:
pairwise Jensen--Shannon divergences between groups; the
dashed line marks their mean, \(\MDS=0.205\).}
\label{fig:site-diagnostics}
\end{figure}

\begin{figure}[!tb]
\centering
\includegraphics[width=\columnwidth]{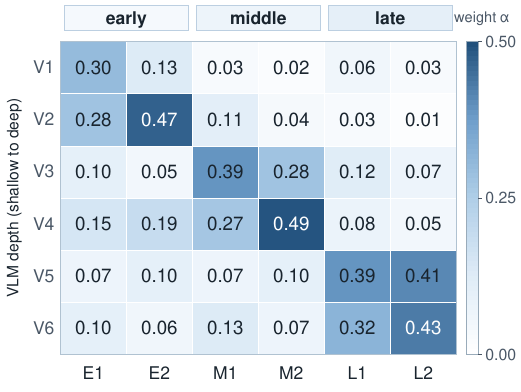}
\caption{layer-wise routing across action-layer groups at 20K.
Columns E1--L2 denote six router bins and rows \(V_1\)--\(V_6\) denote
the six VLM layers. Routing shifts from shallow states in early bins
to middle and deeper states in later bins.}
\label{fig:routing-heatmap}
\end{figure}

\begin{figure*}[!t]
\centering
\includegraphics[width=\textwidth]{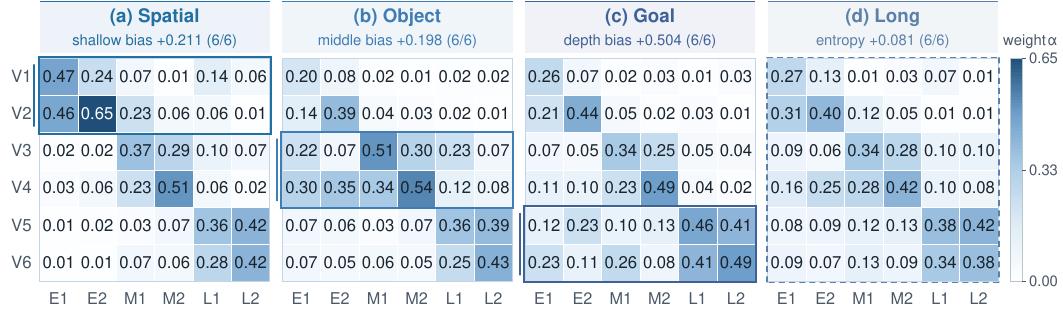}
\caption{Suite-specific layer-wise routing. Columns E1--L2 denote router bins and rows \(V_1\)--\(V_6\) denote VLM layers. Spatial favors shallow states, Object middle states, Goal deeper states, and Long a broader mixture. Subtitles report suite-specific margins relative to the corresponding reference distribution; 6/6 indicates positive margins across all six router bins.}
\label{fig:suite-routing-prior}
\end{figure*}

We analyze aggregate routing behavior across four LIBERO suites
during training. As shown in Figure~\ref{fig:entropy}, the routing
statistics largely stabilize by 20K steps; we therefore use this
checkpoint for the subsequent analyses in
Figures~\ref{fig:site-diagnostics}--\ref{fig:suite-routing-prior}.

To characterize how routing evolves across the action DiT, we analyze
the 18 cross-attention layers where the VLM-depth router is applied.
Following their order in the network, we group consecutive layers into
six bins of three layers each:
\(\mathcal{B}
=
\{\mathrm{E1},\mathrm{E2},\mathrm{M1},
  \mathrm{M2},\mathrm{L1},\mathrm{L2}\}\).
For each bin \(b\in\mathcal{B}\), we first average the routing
distributions of its three layers within each suite and then compute
the equal-weight mean over the four suites, obtaining an aggregate
distribution \(\bar{\boldsymbol{\alpha}}^{(b)}\) over the six VLM
depths. The component \(\bar{\alpha}_n^{(b)}\) denotes the mean routing
weight assigned to VLM depth \(n\) within bin \(b\). We quantify routing
selectivity using the entropy and normalized entropy deficit:
\[
\Hop^{(b)}
=
-\sum_{n=1}^{K}
\bar{\alpha}_{n}^{(b)}
\log \bar{\alpha}_{n}^{(b)},
\quad
\Delta_{\Hop}^{(b)}
=
1-\frac{\Hop^{(b)}}{\log K},
\quad K=6.
\]
Uniform routing gives \(\Delta_{\Hop}^{(b)}=0\), whereas larger values
indicate stronger concentration on fewer VLM layers.

To measure routing variation across action-computation stages, we merge
adjacent bins into early, middle, and late groups:
\(\mathcal{G}_e=\{\mathrm{E1},\mathrm{E2}\}\),
\(\mathcal{G}_m=\{\mathrm{M1},\mathrm{M2}\}\), and
\(\mathcal{G}_l=\{\mathrm{L1},\mathrm{L2}\}\).
For \(g\in\{e,m,l\}\), we average the corresponding bin-level
distributions:
\(\bar{\boldsymbol{\alpha}}_g
=
\frac{1}{2}
\sum_{b\in\mathcal{G}_g}
\bar{\boldsymbol{\alpha}}^{(b)}\).
These group-level distributions characterize the average VLM-depth
allocation at different stages of action computation. We measure their
separation using the mean distribution separation (MDS), computed using Jensen--Shannon divergence:
\[
\MDS
=
\frac{1}{3}
\Bigl[
\JS(\bar{\boldsymbol{\alpha}}_e,\bar{\boldsymbol{\alpha}}_m)
+
\JS(\bar{\boldsymbol{\alpha}}_e,\bar{\boldsymbol{\alpha}}_l)
+
\JS(\bar{\boldsymbol{\alpha}}_m,\bar{\boldsymbol{\alpha}}_l)
\Bigr].
\]
Larger \(\MDS\) indicates greater variation in VLM-depth allocation across action-computation stages,
while \(\MDS=0\) corresponds to identical depth distributions.

\paragraph{Routing becomes more selective during training.}
Figure~\ref{fig:entropy} shows that routing gradually shifts away
from the uniform distribution. The mean normalized entropy
across the six bins decreases from 1.000 at initialization to 0.805 at
20K, while the corresponding normalized entropy deficit increases from
0.000 to 0.195. Both measures stabilize near 20K. These trends indicate
that the router learns increasingly selective VLM-depth allocation
while maintaining a distributed mixture rather than collapsing to a
single depth.

\paragraph{Routing varies across the action DiT.}
The group-averaged normalized entropy deficit increases from 0.126 in
the early layers to 0.205 in the middle layers and 0.253 in the late
layers (Figure~\ref{fig:site-diagnostics}), indicating more selective
depth routing in later action stages. The early--middle, early--late,
and middle--late Jensen--Shannon divergences are 0.163, 0.248, and
0.205, respectively, with the largest separation between the early and
late groups. Their mean yields \(\MDS=0.205\). Figure~\ref{fig:routing-heatmap} reveals a shift in VLM-depth
allocation: early action layers assign more weight to shallower VLM
states, whereas later layers increasingly favor middle and deeper states.

\paragraph{Task suites exhibit distinct layer-routing patterns.}
Figure~\ref{fig:suite-routing-prior} reports routing heatmaps stratified by
task suite. Let \(\bar{\boldsymbol{\alpha}}^{(u,b)}\) denote the routing
distribution for suite \(u\) in router bin \(b\). We define shallow and
middle masses as the summed routing weights over \(V_1\)--\(V_2\) and
\(V_3\)--\(V_4\), and compute the expected layer index as:
\(D^{(u,b)}=\sum_{n=1}^{K} n\,\bar{\alpha}_{n}^{(u,b)}\).

The routing behaviors exhibit clear task-dependent preferences. Spatial tasks
favor shallow layers, particularly in early routing stages, with a
shallow-mass margin of 0.211, reflecting greater reliance on fine-grained
visual and geometric representations. Object tasks maintain higher routing
mass over \(V_3\)--\(V_4\) before gradually shifting toward deeper layers,
yielding a middle-mass margin of 0.198. Goal tasks consistently exhibit the
largest expected layer index across all bins, exceeding the mean of other
suites by 0.504 on average, suggesting stronger reliance on higher-level
semantic representations. Long-horizon tasks maintain the broadest routing
distribution across VLM layers, with normalized entropy higher by 0.081,
indicating that extended tasks benefit from integrating diverse representations across multiple visual, semantic, and action-relevant hierarchical levels.

\section{Real-World Robot Experiments}

\begin{figure}[!tb]
\centering
\includegraphics[width=\columnwidth]{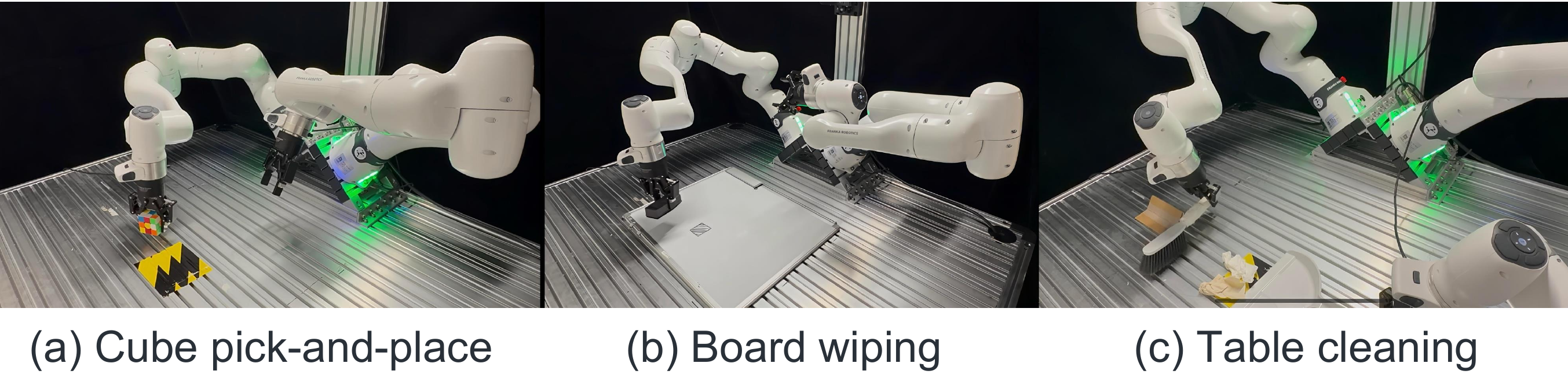}
\caption{Real-world manipulation tasks for evaluating \methodname{} with a
Franka robotic arm. The tasks include (a) cube pick-and-place, (b) eraser-based
board wiping, and (c) broom-based table cleaning, covering diverse grasping,
contact, and tool-use interactions.}
\label{fig:real-world}
\end{figure}

To further evaluate \methodname{} under real-world execution, we conduct
physical robot experiments with a Franka robotic arm equipped with a gripper
across three manipulation tasks. As shown in Figure~\ref{fig:real-world}, the
tasks cover diverse embodied interactions: PnP requires accurate object pose
estimation, stable grasping, and precise placement; Wipe board involves
contact-rich interaction and coordinated tool motion; and Clean Table requires
tool use and longer-horizon object manipulation. Together, these tasks evaluate
whether adaptive representation routing can improve real-world manipulation
performance across different physical interaction patterns.

For each task, we collect 50 demonstration trajectories and fine-tune both
\(\pi_{0.5}\) and \(\pi_{0.5}+\methodname{}\) for 20K steps under identical
training configurations and optimization settings. Each policy is evaluated over 50 independent physical trials per task. As shown in Table~\ref{tab:realworld},
\methodname{} consistently improves task success rates by +8.0, +4.0, and +6.0
percentage points on PnP, Wipe board, and Clean Table, respectively.
Detailed task definitions and training/inference configurations are provided
under \emph{Real-World Experimental Details} in the Supplementary Material.

\begin{table}[!t]
\centering
{\small
\setlength{\tabcolsep}{3pt}
\begin{tabular*}{\linewidth}{@{}l@{\extracolsep{\fill}}ccc@{}}
\toprule
Model & PnP \(\uparrow\) & Wipe board \(\uparrow\) & Clean Table \(\uparrow\) \\
\midrule
\(\pi_{0.5}\) & 72 & 54 & 26 \\
\(\pi_{0.5}\)+\methodname{} & 80 & 58 & 32 \\
\bottomrule
\end{tabular*}
}
\caption{Real-world manipulation success rates (\%) comparing
\(\pi_{0.5}\) and \(\pi_{0.5}+\methodname{}\). \methodname{} consistently improves performance across cube pick-and-place,
board wiping, and table cleaning real-world tasks.}
\label{tab:realworld}
\end{table}

\section{Conclusion}
\label{sec:conclusion}

We identify fixed representation interfaces as a limitation of current VLA
action computation and introduce \methodname{}, an action-conditioned
mixture-of-layers routing framework with adaptive access to VLM and action
representations. By routing over cached VLM layer representations and
revisiting intermediate action states, \methodname{} enables action
computation to dynamically access information beyond fixed layer exposure and
residual-only propagation. Extensive experiments across simulation and
real-world settings demonstrate consistent improvements in VLA
performance with modest overhead. Ablation studies validate the contributions of both representation routing mechanisms, while routing analyses reveal structured
layer allocation patterns that vary across action layers and task settings. These findings highlight adaptive representation routing for flexible robot policies.

\bibliography{references}
\appendix
\section*{Appendix}

\section{Experimental Setup}

This supplement provides implementation details, controlled ablations,
efficiency measurements, and task-level results. Within each backbone,
all component variants share identical training and evaluation settings.
All routed \starvlapi{} controls retain the same \actionreread{} module
and differ only in the VLM-layer routing strategy.

\paragraph{Training data.}
For each backbone, LayerRoute preserves the baseline datasets, mixtures,
preprocessing, normalization, and sampling strategy without adding
demonstrations or task-specific samples. The registered \starvlapi{}
robot-policy mixtures~\citep{community2026starvla} contain 1,693 no-op-filtered LIBERO trajectories
(432 Spatial, 454 Object, 428 Goal, and 379 LIBERO-10), 140,404
SimplerEnv trajectories (53,192 Bridge and 87,212 RT-1/Fractal), and
24,000 RoboCasa-GR1 trajectories (1,000 per task dataset). Each
constituent dataset has registry weight 1.0; these counts exclude the
auxiliary VLM instruction-data stream. The \(\pi_{0.5}\) integration
likewise preserves the original OpenPI~\citep{openpi2026} data configuration. Consequently,
each baseline and LayerRoute pair differs only in its
representation-access interface.

\paragraph{Training configuration.}
All models are trained on eight NVIDIA H200 GPUs with a per-device batch
size of 32. The pretrained VLM, action module, router, and action-state reread
are jointly fine-tuned together with the backbone-specific fusion
parameters. All VLM parameters are updated during fine-tuning. For LIBERO, SimplerEnv, and
RoboCasa-GR1, we use fixed training budgets of 30K, 40K, and 100K
optimization steps, respectively. Both backbones use a 5K-step warmup
during training. We do not perform hyperparameter sweeps. We evaluate the corresponding endpoint checkpoints. For the VLM-backbone sensitivity
experiments, we keep the action architecture and optimization protocol
fixed while replacing the pretrained VLM backbone and jointly
fine-tuning the resulting model under the same configuration.

\paragraph{Randomness control.}
We train three independent runs per benchmark configuration with seeds
\(7\), \(42\), and \(41\), and report the mean performance. For each run,
the global seed is used to initialize Python, NumPy, PyTorch, and CUDA
random number generators before model construction. In distributed
training, each process uses a rank-specific seed derived as \(s+r\),
where \(s\) is the global seed and \(r\) is the process rank. This
ensures deterministic initialization of the router, fusion modules,
and Action-State Reread across distributed workers while preserving
independent stochastic streams. Data-loader workers additionally inherit
rank-specific generator states to ensure reproducible data sampling.

\paragraph{Evaluation checkpoints.}
Figure~2 in the main paper tracks aggregate routing behavior across the
four LIBERO suites during training. The routing statistics largely
stabilize by the 20K-step diagnostic checkpoint, which we use for the
subsequent analyses in Figures~3--5. These routing diagnostics are
separate from the benchmark-specific endpoint evaluations. All
displayed task-level results are obtained by reevaluating the
corresponding models under the same protocols used for their reported
aggregate scores.

\section{Implementation Details}

Table~\ref{tab:supp-impl} summarizes the backbone-specific
implementations. Both implementations preserve the native action
objectives and insert the two representation reads at
backbone-appropriate locations.

\begin{table*}[t]
\centering
{\small
\setlength{\tabcolsep}{1.4pt}
\begin{tabular*}{\textwidth}{@{}>{\raggedright\arraybackslash}p{0.10\textwidth}@{\extracolsep{\fill}}>{\raggedright\arraybackslash}p{0.22\textwidth}>{\raggedright\arraybackslash}p{0.27\textwidth}>{\raggedright\arraybackslash}p{0.34\textwidth}@{}}
\toprule
Backbone & VLM-layer memory & Action interface & Training, checkpoint, and overhead \\
\midrule
\(\pi_{0.5}\) & Online, token-aligned PaliGemma~\citep{beyer2024paligemma} states stored after layers \([2,5,8,11,14,17]\); each site reads only causally available states, with at most six candidates. Here, causal availability refers to layer-wise computation order rather than causal attention masking. & At each of the 18 action-expert layers, an unshared eight-head action-to-VLM cross-attention adapter applies a residual update before native prefix/suffix joint attention. \actionreread{} combines the initial, earlier macro-block, and current action states. A macro-block denotes a retained group of consecutive action-expert layers used as an action-state memory source. & AdamW, learning rate \(5\!\times\!10^{-5}\), a 5K-step linear warmup followed by a constant schedule, and 30K/40K/100K optimization steps on LIBERO/SimplerEnv/RoboCasa-GR1, respectively. Performance is measured using the corresponding endpoint checkpoint. The full interface increases the parameter count by 3.87\% relative to the backbone. \\
\addlinespace[1pt]
\starvlapi{} & Six Qwen3-VL~\citep{bai2025qwen3} states from layers \([5,11,17,23,29,35]\). & At each of the 18 cross-attention layers, a 256-dimensional router provides the routed context directly to native cross-attention. \actionreread{} provides the post-attention reread state directly to the native FFN. & AdamW, action/routing learning rate \(10^{-4}\), a 5K-step warmup followed by cosine decay to \(10^{-7}\), and 30K/40K/100K optimization steps on LIBERO/SimplerEnv/RoboCasa-GR1, respectively. Performance is measured using the corresponding endpoint checkpoint. The full interface increases the parameter count by 0.31\% relative to the backbone. \\
\bottomrule
\end{tabular*}
}
\caption{Backbone-specific implementations of \methodname{}. The
\(\pi_{0.5}\) path~\citep{intelligence2025pi_} reads the selected
memory through a separate action-to-VLM cross-attention adapter before
native joint attention. The \starvlapi{} path~\citep{community2026starvla}
instead supplies the routed memory directly as context to native
cross-attention. Both paths jointly fine-tune the pretrained VLM, action module,
router, action-state reread, and backbone-specific fusion parameters.}
\label{tab:supp-impl}
\end{table*}

\paragraph{Choice of VLM-layer granularity.}
The Layer Mixture Router operates over six cached VLM states in
StarVLA-$\pi$, retained after Qwen3-VL layers
\([5,11,17,23,29,35]\). This granularity follows the block-level state
retention used in Block Attention Residuals~\citep{team2026attention}.
The same six candidates are exposed to all 18 action-DiT
cross-attention routing sites. Grouping these 18 sites into six
consecutive three-site bins is used only for the routing diagnostics
in Figures~3--5 and does not define the VLM candidate set.

We additionally evaluate a denser routing variant with 18 VLM-layer
candidates, where the router directly routes over individual cached
VLM layers. Although this provides finer-grained depth access, we find
that the enlarged routing space leads to less stable router
optimization and does not provide consistent performance gains.
Moreover, the denser variant introduces additional router computation
and inference latency. Therefore, we use the six-depth configuration as
a practical balance between routing granularity, optimization stability,
and computational overhead.

\begin{table}[!t]
\centering
{\small
\setlength{\tabcolsep}{4pt}
\begin{tabular*}{\linewidth}{@{}l@{\extracolsep{\fill}}cc@{}}
\toprule
Routing granularity
& \shortstack{LIBERO\\Avg$_4$ SR (\%) $\uparrow$}
& \shortstack{Latency increase\\(\%) $\downarrow$}
\\
\midrule
6 vlm layers
& \Best{98.0}
& 10.03
\\
18 vlm layers
& 96.2
& 18.47
\\
\bottomrule
\end{tabular*}
}
\caption{Effect of VLM-layer routing granularity on \starvlapi{}.
The 6-depth variant routes over cached Qwen3-VL states from layers
\([5,11,17,23,29,35]\), whereas the 18-depth variant routes over
individual cached VLM layers. All 18 action-DiT cross-attention sites
use the selected candidate set. Latency overhead is measured relative
to the static interface. Bold denotes the higher Avg\(_4\) success rate.}
\label{tab:VLM-layer-granularity}
\end{table}

\paragraph{Optimization details.}
Both implementations use bfloat16 mixed precision and AdamW with
\((\beta_1,\beta_2)=(0.9,0.95)\), \(\epsilon=10^{-8}\), and gradient
clipping at 1.0. The \(\pi_{0.5}\) run uses weight decay
\(10^{-8}\), a per-device batch size of 32, and gradient accumulation
of 1. The \starvlapi{} run uses DeepSpeed ZeRO-2 through Accelerate,
weight decay \(10^{-8}\), and a base-parameter learning rate of
\(2.5\!\times\!10^{-5}\).

\paragraph{Online memory in \(\pi_{0.5}\).}
The configured indices specify where VLM outputs are stored rather than
the candidate states available at each routing site. Before
action-expert layer \(\ell\), the router reads completed boundary
outputs with \(b<\ell\) and additionally reads the layer-\(\ell\) input
when it does not duplicate the immediately preceding boundary. At layer
17, the readable memory bank contains outputs after layers
\(2,5,8,11,14\) and the layer-17 input. The layer-17 output is stored
afterward but is not available to downstream routers. Routing does
not modify VLM execution.

\paragraph{Cross-attention adapter in \(\pi_{0.5}\).}
At zero-based action-expert layer \(s\in\{0,\ldots,17\}\), let
\(\mathbf A_{\mathrm{in}}^{(s)}\in\R^{B\times T_a\times D_a}\) denote
the current action state and let
\(\widetilde{\bV}^{(s)}\in\R^{B\times T_v\times D_v}\) denote the
token-aligned mixture over causally available vlm layers, where
\(D_a=1024\) and \(D_v=2048\). The selected memory is not provided
directly to the native joint-attention interface. Instead, each
action-expert layer applies an unshared action-to-VLM cross-attention
adapter:
\begin{equation}
\begin{aligned}
\mathbf Q^{(s)}
&=\mathbf A_{\mathrm{in}}^{(s)}
  \left(\mathbf W_q^{(s)}\right)^\top,\\
\mathbf K^{(s)}
&=\widetilde{\bV}^{(s)}
  \left(\mathbf W_k^{(s)}\right)^\top,\\
\mathbf U^{(s)}
&=\widetilde{\bV}^{(s)}
  \left(\mathbf W_v^{(s)}\right)^\top,\\
\mathbf S_h^{(s)}
&=\frac{\mathbf Q_h^{(s)}
  \left(\mathbf K_h^{(s)}\right)^\top}{\sqrt{d_h}}
  +\mathcal M,\\
\mathbf O_h^{(s)}
&=\softmax\!\left(\mathbf S_h^{(s)}\right)\mathbf U_h^{(s)},\\
\mathbf C^{(s)}
&=\operatorname{Concat}_{h=1}^{H}\!\left(\mathbf O_h^{(s)}\right)
  \left(\mathbf W_o^{(s)}\right)^\top,\\
\mathbf A_{\mathrm{pre}}^{(s)}
&=\mathbf A_{\mathrm{in}}^{(s)}
  +\mathbf C^{(s)}.
\end{aligned}
\label{eq:supp-pi05-cross-attn}
\end{equation}
The adapter uses \(H=8\) heads with
\(d_h=D_a/H=128\). Its bias-free matrices have dimensions
\(\mathbf W_q^{(s)},\mathbf W_o^{(s)}\in\R^{D_a\times D_a}\) and
\(\mathbf W_k^{(s)},\mathbf W_v^{(s)}\in\R^{D_a\times D_v}\);
\(\mathcal M\) excludes invalid VLM tokens. The attention is
non-causal and uses zero dropout. The four matrices are initialized
with the default PyTorch linear-layer scheme,
\(\operatorname{KaimingUniform}(a=\sqrt{5})\), and do not use biases.
The router's action-side depth-query projection is zero-initialized,
so the initial depth distribution is uniform over the available causal
states. All adapter parameters are optimized during fine-tuning.

The updated state \(\mathbf A_{\mathrm{pre}}^{(s)}\) becomes the
action-suffix input to native prefix/suffix joint attention, while the
adapter leaves the VLM-prefix input unchanged at this insertion point.
After native attention, \actionreread{} is applied before the action
FFN. Training and inference follow the same execution order. Together,
the Q/K/V/O projections, per-site routers, normalization layers, and
\actionreread{} increase the parameter count by 3.87\% relative to the
backbone.

\paragraph{router and action-state reread.}
The routers pool action and VLM tokens, apply separate LayerNorm
operations, and use site-specific query and key projections. Both full
and router-only variants use the current action state
\(\mathbf a^{(s)}\) to construct the router query. Only the
\starvlapi{} initial-state control uses the pass-level state
\(\mathbf a_{\mathrm{init}}\) for depth selection (pass-level state refers to the incoming action state at the beginning of the forward/denoising pass); the current action
state at each site remains the input to native cross-attention and its
residual stream. Each router predicts a single sample- and site-specific
distribution shared across token positions, whereas
\actionreread{} produces token-specific weights. Neither read uses an
auxiliary loss or an additional expert.

For \starvlapi{}, let \(\widetilde{\bV}^{(s)}\) denote the routed VLM
representation and let \(\widehat{\mathbf a}^{(s)}\) denote the
token-wise reread action state. The outputs supplied to downstream
sublayers are:
\[
\begin{aligned}
\mathbf Z^{(s)}
&=\widetilde{\bV}^{(s)},\\
\mathbf a_{\mathrm{out}}^{(s)}
&=\widehat{\mathbf a}^{(s)}.
\end{aligned}
\]

Both backbones apply this action-state assignment at every configured
reread site. In \starvlapi{}, \(\mathbf Z^{(s)}\) replaces the original
encoder context, and \(\mathbf a_{\mathrm{out}}^{(s)}\) is passed to the
native FFN. In \(\pi_{0.5}\), the cross-attention adapter first produces
\(\mathbf A_{\mathrm{pre}}^{(s)}\); after native joint attention,
\(\mathbf a_{\mathrm{out}}^{(s)}\) is likewise passed to the native
action FFN.

\paragraph{Action-state memory locations.}
All layer indices are zero-based. In \starvlapi{}, post-FFN-residual
action states are retained after the 18 odd DiT blocks
\(\{1,3,\ldots,35\}\). The pre-FFN reread at each block \(s\) accesses
detached copies of the initial state, all retained states with indices
below \(s\), and the current post-attention-residual state.

In \(\pi_{0.5}\), post-FFN-residual action states are retained
after layers \(\{2,5,8,11,14,17\}\). The pre-FFN reread at each layer
\(s\) accesses the initial state, all retained states with indices below
\(s\), and the current post-attention-residual state.

\section{Compute Environment}

Table~\ref{tab:supp-environment} summarizes the hardware and software
configurations used in our experiments. The two backbone
implementations and simulator clients run in isolated environments and
communicate through the existing policy server, preserving their native
training stacks without modifying optimization or evaluation protocols.

\begin{table*}[t]
\centering
{\small
\setlength{\tabcolsep}{2.2pt}
\begin{tabular*}{\textwidth}{@{}>{\raggedright\arraybackslash}p{0.09\textwidth}@{\extracolsep{\fill}}>{\raggedright\arraybackslash}p{0.13\textwidth}>{\raggedright\arraybackslash}p{0.71\textwidth}@{}}
\toprule
Category & Component & Specification \\
\midrule
Hardware & GPU & 8\(\times\) NVIDIA H200 (Hopper, compute capability 9.0), 143,771 MiB (about 140.4 GiB; nominal 141 GB HBM3e) per GPU and about 1.10 TiB total; 700 W power limit per GPU; NV18 NVLink/NVSwitch all-to-all interconnect; MIG disabled. \\
& CPU & 2\(\times\) Intel Xeon Platinum 8468; 96 physical cores and 192 hardware threads across two NUMA nodes; cgroup quota equivalent to 144 logical CPUs. \\
& Host memory & 1,600 GiB container limit; about 2.0 TiB physically visible; no swap. \\
\addlinespace[1pt]
Execution & Policy stacks & StarVLA and \(\pi_{0.5}\) use separate isolated environments with their native dependencies. \\
& StarVLA stack & Linux x86\_64 (containerized); Python 3.11; CUDA runtime 12.4.127; PyTorch 2.6.0+cu124; torchvision 0.21.0+cu124; Transformers 4.57.0; Accelerate 1.13.0; DeepSpeed 0.16.9; NumPy 1.26.4. The release pins this stack in \texttt{environments/starvla-training.lock.txt}. \\
& \(\pi_{0.5}\) stack & Linux x86\_64; Python 3.11; PyTorch 2.7.1; JAX/jaxlib 0.5.3; Flax 0.10.2; Transformers 4.53.2; NumPy 1.26.4. The authoritative release lock is \texttt{pi05/openpi/uv.lock}. \\
& Simulators & SimplerEnv commit \texttt{06accaca93535902d408da4855f21cece12bceb7} (upstream Conda environment); RoboCasa-GR1 commit \texttt{4840e671596f93ca03651524b9f72ffb1aadfeff} with NumPy 1.26.4, Numba 0.61.2, MuJoCo 3.2.6, and Tianshou 0.5.1. JSON locks record both evaluator environments. The released runtime verifier records the Linux distribution, NVIDIA driver, CUDA runtime, NCCL, and installed packages per run. \\
\bottomrule
\end{tabular*}
}
\caption{Hardware and software used for the controlled experiments.
For the robot-policy training stream, both integrations use a per-device batch size of 32 and gradient
accumulation of 1, yielding a global batch size of 256.}
\label{tab:supp-environment}
\end{table*}

\paragraph{Cross-benchmark execution.}
For LIBERO endpoint evaluation, each model is evaluated on 50 episodes
per task across all 40 tasks (2,000 episodes per model), using base
evaluation seed 7.

For SimplerEnv~\citep{li2024evaluating}, both integrations are trained
on the same LeRobot conversions of Bridge~\citep{ebert2021bridge} and
RT-1~\citep{brohan2022rt}, with equal source probability and
benchmark-specific normalization. The evaluator preserves the public
visual-matching, variant-aggregation, and WidowX protocols; only the
policy call is replaced by WebSocket communication. Following the
official \starvlapi{} protocol~\citep{community2026starvla}, we run the complete
evaluator five times per reported setting and report the mean success
rate. All five repetitions use the same endpoint checkpoint and do not
constitute independent training runs. Each repetition uses 24 episodes
per WidowX task.

For Google Robot, the per-repetition episode counts for Coke, Near,
Drawer, and Put-in are 300/240/216/108 under VM and
825/600/378/189 under VA, giving five-repetition totals of
1,500/1,200/1,080/540 and 4,125/3,000/1,890/945, respectively. Each
repetition follows the same benchmark-defined episode grid; no
additional environment-seed averaging is applied.

For RoboCasa-GR1, both integrations use the same 24-dataset mixture,
equal source probability, and the same 24-task evaluator with 50
rollouts per task (1,200 rollouts per model), without additional
environment-seed averaging. The \starvlapi{} path applies its
registered 58-dimensional sine/cosine state transform. The
\(\pi_{0.5}\) path concatenates the corresponding 29 raw joint values,
discretizes this state into prefix tokens, and pads only the action
target to its internal 32-dimensional width. The released code provides
the training, normalization, policy-serving, and evaluation commands in
the SimplerEnv, RoboCasa-GR1, and \(\pi_{0.5}\) code directories.

\begin{table*}[t]
\centering
{\small
\setlength{\tabcolsep}{3.5pt}

\begin{tabular*}{\textwidth}{
@{}>{\raggedright\arraybackslash}p{0.52\textwidth}
@{\extracolsep{\fill}}
>{\centering\arraybackslash}p{0.095\textwidth}
>{\centering\arraybackslash}p{0.095\textwidth}
>{\centering\arraybackslash}p{0.115\textwidth}
>{\centering\arraybackslash}p{0.115\textwidth}@{}}
\toprule
&
\multicolumn{4}{c}{Success rate (\%) \(\uparrow\)} \\
\cmidrule(l){2-5}
Task
& \(\pi_{0.5}\)
& \starvlapi{}
& \shortstack{\(\pi_{0.5}\)\\+\methodname{}}
& \shortstack{\starvlapi{}\\+\methodname{}} \\
\midrule

PnP Bottle To Cabinet Close
& 20 & 26 & 26 & 30 \\

PnP Can To Drawer Close
& 54 & 62 & 58 & 62 \\

PnP Cup To Drawer Close
& 34 & 42 & 42 & 44 \\

PnP Milk To Microwave Close
& 44 & 50 & 48 & 52 \\

PnP Potato To Microwave Close
& 32 & 42 & 42 & 44 \\

PnP Wine To Cabinet Close
& 26 & 32 & 32 & 36 \\

PnP Novel From Cuttingboard To Basket
& 36 & 40 & 38 & 42 \\

PnP Novel From Cuttingboard To Cardboardbox
& 38 & 46 & 46 & 48 \\

PnP Novel From Cuttingboard To Pan
& 52 & 60 & 50 & 58 \\

PnP Novel From Cuttingboard To Pot
& 32 & 40 & 40 & 42 \\

PnP Novel From Cuttingboard To Tieredbasket
& 36 & 44 & 42 & 44 \\

PnP Novel From Placemat To Basket
& 38 & 44 & 42 & 46 \\

PnP Novel From Placemat To Bowl
& 46 & 52 & 50 & 52 \\

PnP Novel From Placemat To Plate
& 42 & 50 & 48 & 50 \\

PnP Novel From Placemat To Tieredshelf
& 22 & 28 & 32 & 32 \\

PnP Novel From Plate To Bowl
& 46 & 52 & 50 & 52 \\

PnP Novel From Plate To Cardboardbox
& 34 & 40 & 40 & 40 \\

PnP Novel From Plate To Pan
& 30 & 36 & 38 & 38 \\

PnP Novel From Plate To Plate
& 40 & 48 & 44 & 48 \\

PnP Novel From Tray To Cardboardbox
& 28 & 34 & 36 & 36 \\

PnP Novel From Tray To Plate
& 56 & 64 & 56 & 62 \\

PnP Novel From Tray To Pot
& 36 & 44 & 42 & 44 \\

PnP Novel From Tray To Tieredbasket
& 44 & 50 & 48 & 50 \\

PnP Novel From Tray To Tieredshelf
& 22 & 28 & 30 & 30 \\

\midrule
Average
& 37.0
& 43.9
& 42.5
& \Best{45.1} \\
\bottomrule
\end{tabular*}
}
\caption{RoboCasa-GR1 task-level success rates (\%) across 24
tasks~\citep{nasiriany2024robocasa}, obtained from our
reevaluation under the common benchmark protocol. The Average row
reports the macro-average across all 24 tasks. Bold denotes the highest
macro-average.}
\label{tab:supp-robocasa-task}
\end{table*}

\section{Component Ablations}

Table~\ref{tab:supp-routing-controls} reports additional routing and
component controls for \starvlapi{}. All variants are fine-tuned with
the same optimization protocol, including updates to the VLM parameters,
and use the same data, optimization budget, and evaluation protocol.

\begin{table}[!b]
\centering
{\small
\setlength{\tabcolsep}{1.2pt}
\begin{tabular*}{\linewidth}{@{}>{\raggedright\arraybackslash}p{0.49\linewidth}@{\extracolsep{\fill}}ccc@{}}
\toprule
Routing configuration & \shortstack{Avg\(_4\)\\SR (\%) \(\uparrow\)} & \shortstack{Long\\SR (\%) \(\uparrow\)} & \shortstack{Long gain\\(pp) \(\uparrow\)} \\
\midrule
original static interface & 95.7 & 88.4 & 0.0 \\
\midrule
static mixture (no action reread) & 95.6 & 90.0 & +1.6 \\
initial-state route & 96.4 & 90.8 & +2.4 \\
current-state route & \Best{98.0} & \Best{95.6} & \Best{+7.2} \\
\bottomrule
\end{tabular*}
}
\caption{Routing and component controls for \starvlapi{}.
Long gain is measured in percentage points relative to the original
static interface. The initial-state and current-state routes differ only in
the router query state and retain the same \actionreread{} module.
The static-mixture row disables \actionreread{} and therefore does
not isolate routing policy. Bold denotes the best result in each metric.}
\label{tab:supp-routing-controls}
\end{table}

\section{Ablation Details}

For \starvlapi{}, let \(P_\ell\), \(\ell=0,\ldots,17\), denote the
router at the \(\ell\)-th DiT cross-attention site, corresponding to
zero-based DiT blocks \(\{0,2,\ldots,34\}\). At the beginning of each
training forward pass or inference denoising pass, \(P_0\) captures its
incoming action state without detachment:
\[
\mathbf a_{\mathrm{init}}
=
\mathbf a_{\mathrm{in}}^{(P_0)}.
\]

Let \(\bV_n\), \(n\in\mathcal C=\{1,\ldots,6\}\), denote the six
token-aligned VLM states cached after Qwen3-VL layers
\([5,11,17,23,29,35]\), and let
\(\mathbf p_{V,n}=\Pool(\bV_n)\) denote the corresponding mask-aware
pooled summary. Every router uses the same candidate set
\(\mathcal C_s=\mathcal C\). The following definitions cover the
controls in Table~\ref{tab:supp-routing-controls} and Table~7 of the
main paper, using router dimension \(d_{\mathrm{sel}}=256\):
\[
\begin{aligned}
\mathbf Z_{\mathrm{static}}^{(s)}
&=\Phi_s\!\left(\{\bV_n\}_{n\in\mathcal C_s}\right),\\
\widetilde{\bV}_{\mathrm{mix}}^{(s)}
&=\sum_{n\in\mathcal C_s}
  \softmax\!\left(\boldsymbol{\beta}^{(s)}\right)_n\bV_n,\\
\bq_{\mathrm{ind}}^{(s)}
&=\mathbf W_{q,d}^{(s)}
  \layernorm_{a}^{(s)}\!\left(\mathbf c_0\right),
  \qquad \mathbf c_0\ne\mathbf 0,\\
\mathbf k_n^{(s)}
&=\mathbf W_{k,d}^{(s)}
  \layernorm_{V}^{(s)}\!\left(\mathbf p_{V,n}\right),\\
\ell_{n,\mathrm{ind}}^{(s)}
&=\frac{1}{\sqrt{d_{\mathrm{sel}}}}
  \langle
  \bq_{\mathrm{ind}}^{(s)},\mathbf k_n^{(s)}
  \rangle,
  \qquad n\in\mathcal C_s,\\
\mathbf p_{a,\mathrm{init}}
&=\Pool\!\left(\mathbf a_{\mathrm{init}}\right),\\
\bq_{\mathrm{init}}^{(s)}
&=\mathbf W_{q,d}^{(s)}
  \layernorm_{a}^{(s)}\!\left(\mathbf p_{a,\mathrm{init}}\right),\\
\ell_{n,\mathrm{init}}^{(s)}
&=\frac{1}{\sqrt{d_{\mathrm{sel}}}}
  \langle
  \bq_{\mathrm{init}}^{(s)},\mathbf k_n^{(s)}
  \rangle,
  \qquad n\in\mathcal C_s,\\
\alpha_{n,\mathrm{init}}^{(s)}
&=
  \frac{\exp\!\left(\ell_{n,\rm init}^{(s)}\right)}
       {\sum_{m\in\mathcal C_s}
        \exp\!\left(\ell_{m,\rm init}^{(s)}\right)},\\
\widetilde{\bV}_{\rm init}^{(s)}
&=\sum_{n\in\mathcal C_s}
  \alpha_{n,\rm init}^{(s)}\bV_n.
\end{aligned}
\]
The native static interface preserves each backbone's
architecture-fixed VLM pathway and is independent of the current action
state. The reported \starvlapi{} static-mixture control learns one set
of action-independent logits \(\boldsymbol{\beta}^{(s)}\) per site and
disables \actionreread{}. The action-independent query control retains
the learned site-specific projections and content-dependent keys, but
replaces the pooled action representation with the fixed nonzero vector
\(\mathbf c_0\), as defined in the main paper.

The \starvlapi{} initial-state route uses
\(\mathbf a_{\mathrm{init}}\) only to construct the router query. Each
site retains the same six VLM candidates and site-specific projections;
the current action state remains the input to native cross-attention and
its residual stream. The initial-state and current-state routes use the
same \actionreread{} configuration.

\section{Routing-Policy Controls}

The \(\Delta_{\mathrm{Long}}\) column in
Table~\ref{tab:supp-routing-controls} uses the original static interface
as the reference. For \starvlapi{}, the initial-state route conditions
every router on the action state entering the first router site,
whereas the current-state route uses the action state entering the
corresponding cross-attention site. Both routes use the same six-state
VLM memory bank and \actionreread{}, thereby isolating the effect of the
router query state.

The reported \starvlapi{} static-mixture row also disables
\actionreread{}. Therefore, its comparison with the current-state route
combines the effects of routing dependence and action-state rereading,
and is interpreted only as a component comparison. For \starvlapi{},
all routed variants in Table~7 of the main paper retain the same
action-state reread module and differ only in the VLM-layer routing
rule.

Uniform routing assigns equal weight to all six vlm layers. The shallow,
middle, and deep routes use the
\(\{\bV_1,\bV_2\}\), \(\{\bV_3,\bV_4\}\), and
\(\{\bV_5,\bV_6\}\) groups defined in the main paper. The
action-independent query preserves the router parameterization while
removing dependence on the current action state. The current-state route
outperforms this control by 0.8 points on both Avg\(_4\) and Long.
Comparisons among routed variants in Table~7 of the main paper isolate
the routing rule, whereas comparisons with the original interface
additionally include the effect of action-state rereading. In
Table~\ref{tab:supp-routing-controls}, only the initial-state versus
current-state comparison controls for the same \actionreread{}
configuration.

\begin{figure*}[!t]
\centering
\includegraphics[width=\textwidth]{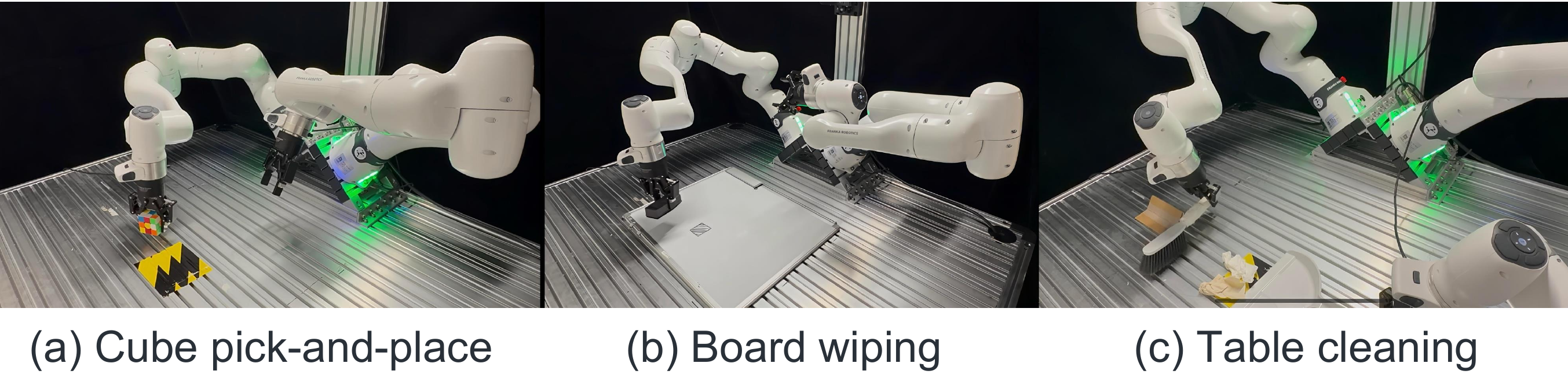}
\caption{
Real-world manipulation task setups.
(a) PnP evaluates precise object localization, grasping, and placement.
(b) Wipe board involves contact-rich tool interaction and spatial coverage.
(c) Clean Table requires longer-horizon tool-use manipulation in a cluttered
workspace.
}
\label{fig:realworld-task}
\end{figure*}

\subsection{Routing Diagnostics}

Figure~2 in the main paper tracks aggregate routing behavior across the
four LIBERO~\citep{liu2023libero} suites during training. The routing
statistics largely stabilize by the 20K-step diagnostic checkpoint,
which is used for the subsequent analyses in Figures~3--5. These
diagnostics are separate from the benchmark-specific endpoint
checkpoints used for performance evaluation.

We group the 18 cross-attention router sites by zero-based DiT index:
\[
\begin{aligned}
\mathrm{E1}&=\{0,2,4\}, &
\mathrm{E2}&=\{6,8,10\},\\
\mathrm{M1}&=\{12,14,16\}, &
\mathrm{M2}&=\{18,20,22\},\\
\mathrm{L1}&=\{24,26,28\}, &
\mathrm{L2}&=\{30,32,34\}.
\end{aligned}
\]
Each profile is the arithmetic mean of its three router
distributions. For each router bin, we first average routing distributions across layers and suites, then normalize the resulting distribution over vlm layers. These analyses characterize
aggregate routing behavior and support suite-level interpretations of
how VLM-layer allocation relates to the visual, semantic, and
action-relevant demands of different task suites; they do not attribute
individual rollout failures to a specific vlm layer or routing decision.

\section{Additional Experimental Results}

\subsection{Reporting Conventions and Data Sources}

\begin{table}[!b]
\centering
{\small
\setlength{\tabcolsep}{2.0pt}
\begin{tabular*}{\linewidth}{@{}>{\raggedright\arraybackslash}p{0.22\linewidth}@{\extracolsep{\fill}}>{\raggedright\arraybackslash}p{0.24\linewidth}>{\centering\arraybackslash}p{0.20\linewidth}>{\centering\arraybackslash}p{0.26\linewidth}@{}}
\toprule
Backbone & Interface variant & \shortstack{Added\\parameters\\(\%) \(\downarrow\)} &
\shortstack{Inference latency\\increase (\%) \(\downarrow\)} \\
\midrule
\(\pi_{0.5}\) & Static & 0.00\% & 0.00\% \\
\(\pi_{0.5}\) & router-only & +3.45\% & +14.57\% \\
\(\pi_{0.5}\) & Full & +3.87\% & +25.04\% \\
\midrule
\starvlapi{} & Static & 0.00\% & 0.00\% \\
\starvlapi{} & router-only & +0.27\% & +7.42\% \\
\starvlapi{} & Full & +0.31\% & +10.03\% \\
\bottomrule
\end{tabular*}
}
\caption{Parameter and inference-latency overheads. All percentages
denote increases relative to the corresponding original static
interface.}
\label{tab:supp-efficiency-context}
\end{table}

\begin{table*}[!t]
\begin{minipage}[t]{0.455\textwidth}
\vspace{0pt}
\centering
{\small
\setlength{\tabcolsep}{0.4pt}
\begin{tabular*}{\linewidth}{@{}>{\raggedright\arraybackslash}p{0.42\linewidth}@{\extracolsep{\fill}}ccccc@{}}
\toprule
Model & Spoon & Carrot & Stack & \shortstack{Egg-\\plant} & \shortstack{Pooled\\SR} \\
\midrule
\mbox{\(\pi_{0.5}\) (original)}
& 49.2 & 65.0 & 44.2 & 70.0 & 57.1 \\

\mbox{\starvlapi{}}
& 78.3 & 46.7 & 30.0 & 88.3 & 60.8 \\

\mbox{\(\pi_{0.5}\)+\methodname{}}
& 59.2 & 68.3 & 54.2 & 69.2 & 62.7 \\

\mbox{\starvlapi{}+\methodname{}}
& 79.2 & 50.0 & 36.7 & 87.5 & \Best{63.3} \\
\bottomrule
\end{tabular*}
}
\caption{WidowX Visual Matching task success rates (\%), obtained from
our reevaluation under the common benchmark protocol. Pooled SR is
computed from pooled success counts before task-level rates are rounded.
Bold denotes the highest pooled SR.}
\label{tab:supp-simpler-widowx-detail}
\end{minipage}
\hfill
\begin{minipage}[t]{0.537\textwidth}
\vspace{0pt}
\centering
{\small
\setlength{\tabcolsep}{0.4pt}
\begin{tabular*}{\linewidth}{@{}>{\raggedright\arraybackslash}p{0.37\linewidth}@{\extracolsep{\fill}}ccccc@{}}
\toprule
Model & \shortstack{Coke\\VM/VA} & \shortstack{Near\\VM/VA} & \shortstack{Drawer\\VM/VA} & \shortstack{Put-in\\VM/VA} & \shortstack{Overall\\SR} \\
\midrule
\mbox{\(\pi_{0.5}\)}
& 91.8/88.6 & 72.4/71.1 & 66.2/52.9 & 64.4/57.4 & 70.6 \\

\mbox{\starvlapi{}}
& 94.1/90.1 & 73.8/73.9 & 67.6/53.8 & 65.0/58.2 & 72.1 \\

\mbox{\(\pi_{0.5}\)+\methodname{}}
& 93.2/89.6 & 75.0/72.9 & 69.3/56.7 & 66.9/58.0 & 72.7 \\

\mbox{\starvlapi{}+\methodname{}}
& 94.9/91.3 & 75.5/74.8 & 70.1/56.6 & 66.9/59.6 &
\Best{73.7} \\
\bottomrule
\end{tabular*}
}
\caption{Google Robot task-family success rates (\%) from our
common-protocol reevaluation. VM/VA denote Visual Matching/Variant
Aggregation. Aggregate VM and VA are unweighted means across the four
task families; Overall SR is their mean. Bold denotes the highest
Overall SR.}
\label{tab:supp-simpler-google-detail}
\end{minipage}
\end{table*}

\paragraph{Evaluation metrics.}
We use success rate (SR) as the primary metric for all manipulation
benchmarks, where a rollout is counted as successful if the task-specific
completion criterion is satisfied. For \(N_{\mathrm{succ}}\) successful
rollouts among \(N_{\mathrm{eval}}\) evaluated rollouts, we compute
\(\mathrm{SR}=100N_{\mathrm{succ}}/N_{\mathrm{eval}}\). SR is selected because it directly
measures whether the learned policy completes the target manipulation
objective, which is the standard evaluation criterion for robot control
benchmarks.

For LIBERO, we report the success rate of each task suite and the
unweighted macro-average across the four suites, denoted as Avg\(_4\),
where \(\mathrm{Avg}_4=\frac{1}{4}\sum_{j=1}^{4}\mathrm{SR}_j\). For
SimplerEnv, we report the benchmark-defined Overall success rate together
with the WidowX success rate. For RoboCasa-GR1, we report the
macro-average success rate over the 24 evaluated tasks, treating each
task equally regardless of rollout count. Higher values indicate better
manipulation performance. Within each backbone, component variants use identical
training and evaluation settings.

Table~\ref{tab:supp-efficiency-context} reports parameter and
inference-latency overheads for both backbones. The additional latency
mainly comes from the router computation and the attention operations
introduced by the routing and reread modules. Latency is measured on the
same hardware for all variants with batch size one. We discard the first
50 warm-up inference runs and report the average latency over the
following 100 inference runs. The parameter-count and latency overheads depend in part on the retained vlm layers and action-state locations, which remain fixed for each backbone.

All entries in Tables~\ref{tab:supp-simpler-widowx-detail} and
\ref{tab:supp-simpler-google-detail} are obtained through reevaluation
of the displayed models. Within each benchmark, every model is evaluated
under the same protocol used for its corresponding aggregate score.

\section{Real-World Experimental Details}
\label{sec:supp-realworld}

\paragraph{Robot setup.}
We conduct real-world experiments using a Franka robotic arm equipped with a
parallel gripper. The policy receives RGB observations and robot
proprioceptive states as inputs, following the same observation and action
interfaces used during training. RGB observations are captured using
SC3000 mounted around the workspace, and all experiments are conducted
in a fixed tabletop manipulation environment. The control frequency is set to
10 Hz, and the workspace configuration remains unchanged across all trials to
ensure consistent evaluation.

\paragraph{Task setup.}
We evaluate \methodname{} on three real-world manipulation tasks:
Pick-and-Place (PnP), Wipe board, and Clean Table
(Figure~\ref{fig:realworld-task}). These tasks cover complementary
manipulation requirements, including object grasping and placement,
contact-rich surface interaction, and longer-horizon tool-use coordination.
For each task, we collect 50 demonstration trajectories and fine-tune both
\(\pi_{0.5}\) and \(\pi_{0.5}+\methodname{}\) for 20K steps under identical
training configurations.

\paragraph{Evaluation protocol.}
All models are evaluated using the same real-world protocol without additional
fine-tuning. For each task, we perform 50 independent evaluation trials using
the trained checkpoint. Task success is determined according to task-specific
completion criteria: successful object placement for PnP, successful board
wiping for Wipe board, and successful table cleaning for Clean Table. Reported
success rates are computed over the 50 trials for each task. Detailed hardware
specifications, task definitions, training and inference configurations, and
evaluation criteria are provided in the Supplementary Material.

\section{Limitations}

\methodname{} introduces additional parameter and inference-latency
overheads in both backbones, which may constrain deployment in
resource-limited settings. We evaluate \methodname{} on only
\starvlapi{} and \(\pi_{0.5}\); extending the evaluation to broader VLA
architectures and reducing overhead remain important directions for
future work. In addition, LayerRoute relies on access to intermediate VLM and
action states, which may require architecture-specific integration for
new VLA designs.
Layeroute
\end{document}